\documentclass[10pt,journal,compsoc]{IEEEtran}

\newcommand{\ie}{\textit{i}.\textit{e}.}
\newcommand{\eg}{\textit{e}.\textit{g}.}

\usepackage{hyperref}       
\hypersetup{
colorlinks=true,
linkcolor=red,
anchorcolor=green,
citecolor=green}

\usepackage{bbm}
\usepackage{pifont}
\usepackage{xcolor}
\usepackage{amssymb}

\definecolor{mygray}{gray}{0.4}

\ifCLASSOPTIONcompsoc
  \usepackage[nocompress]{cite}
 \usepackage{amsmath,amsfonts}
  \usepackage{algorithmic}
  \usepackage{algorithm}
  \usepackage{array}
  \usepackage{textcomp}
  \usepackage{stfloats}
  \usepackage{url}
  \usepackage{verbatim}
  \usepackage{graphicx}
  \usepackage{cite}
  \usepackage{multirow}
  \usepackage{amsthm}
  \usepackage{booktabs}
  \usepackage{colortbl}
  \usepackage{caption}
  \usepackage{multirow}
\usepackage{booktabs}
  \usepackage{subcaption}
  
\else
  \usepackage{cite}
\fi

\ifCLASSINFOpdf
\else
\fi

\usepackage{amsmath}
\usepackage{array}
\usepackage{pifont}

\usepackage{fixltx2e}
\usepackage{stfloats}
\usepackage{tikz}
\newcommand*{\circled}[1]{\lower.7ex\hbox{\tikz\draw (0pt, 0pt)%
    circle (.4em) node {\makebox[1em][c]{\small #1}};}}

\usepackage{url}

\begin{document}

\title{EgoMotion: Hierarchical Reasoning and Diffusion for Egocentric Vision-Language Motion Generation}

\author{Ruibing~Hou, 
        Mingyue~Zhou,
        Yuwei~Gui,
        Mingshuang~Luo, 
        Bingpeng~Ma,
        Hong~Chang, 
        Shiguang~Shan~\IEEEmembership{Fellow,~IEEE}
        and~Xilin~Chen~\IEEEmembership{Fellow,~IEEE}
\IEEEcompsocitemizethanks{\IEEEcompsocthanksitem R. Hou and M Luo are with Key Laboratory of Intelligent Information Processing, Institute of Computing Technology (ICT), Chinese Academy of Sciences (CAS), Beijing, 100190, China. 
\IEEEcompsocthanksitem M. Zhou is with Jilin University (JLU), Changchun, 130012, China.
\IEEEcompsocthanksitem Y. Gui is  with Beijing University of Posts and Telecommunications (BUPT), Beijing, 100876, China.
\IEEEcompsocthanksitem  B. Ma is with University of the Chinese Academy of Sciences, Beijing 100049, China.
 \IEEEcompsocthanksitem  H. Chang, S. Shan and X. Chen are with Key Laboratory of Intelligent Information Processing, Institute of Computing Technology (ICT), Chinese Academy of Sciences (CAS), Beijing, 100190, China, and University of the Chinese Academy of Sciences, Beijing 100049, China.
 \protect\\
E-mail: \{houruibing, changhong, sgshan\}@ict.ac.cn, mingshuang.luo@vipl.ict.ac.cn}

}

\markboth{Journal of \LaTeX\ Class Files,~Vol.~14, No.~8, August~2015}%
{Shell \MakeLowercase{\textit{et al.}}: Bare Demo of IEEEtran.cls for Computer Society Journals}

\IEEEtitleabstractindextext{%
\begin{abstract}
Faithfully modeling human behavior in dynamic environments is a foundational challenge for embodied intelligence. While conditional motion synthesis has achieved significant advances, egocentric motion generation remains largely underexplored due to the inherent complexity of first-person perception. In this work, we investigate Egocentric
Vision-Language (Ego-VL) motion generation. This task requires synthesizing 3D human motion conditioned jointly on first-person visual observations and natural language instructions. We identify a critical \textit{reasoning-generation entanglement} challenge: the simultaneous optimization of semantic reasoning and kinematic modeling introduces gradient conflicts. These conflicts systematically degrade the fidelity of multimodal grounding and motion quality. To address this challenge, we propose a hierarchical generative framework \textbf{EgoMotion}. Inspired by the biological decoupling of cognitive reasoning and motor control, EgoMotion operates in two stages. 
In the Cognitive Reasoning stage, A vision-language model (VLM) projects multimodal inputs into a structured space of discrete motion primitives. This forces the VLM to acquire goal-consistent representations, effectively bridging the semantic gap between high-level perceptual understanding and low-level action execution. 
In the Motion Generation stage, these learned representations serve as expressive conditioning signals for a diffusion-based motion generator. By performing iterative denoising within a continuous latent space, the generator synthesizes physically plausible and temporally coherent trajectories.  
Extensive evaluations demonstrate that EgoMotion achieves state-of-the-art performance, and produces motion sequences that are both semantically grounded and kinematically superior to existing approaches.  

\end{abstract}

}

\maketitle

\IEEEdisplaynontitleabstractindextext
\IEEEpeerreviewmaketitle

\IEEEraisesectionheading{\section{Introduction}\label{sec:introduction}}
Faithfully modeling human behavior in dynamic environments remains a foundational challenge with broad implications for computer vision \cite{tanaka2023role,liang2024intergen,jiang2023motiongpt,tian2023recovering,xu20213d}, computer graphics \cite{jiang2024autonomous}, and robotics \cite{lee2023locomotion}. Recent advances in conditional motion generation have largely converged around tasks such as text-to-motion, music-to-dance and scene-conditioned motion synthesis, giving rise to three dominant paradigms: (i) \textit{autoregressive models}, which formulate motion synthesis as sequential token prediction \cite{jiang2023motiongpt,zhang2024motiongpt}; (ii) \textit{masked transformers}, which enable parallel generation through discrete token reconstruction \cite{guo2024momask}; 
(iii) \textit{diffusion-based models}, which model complex, continuous motion distributions through iterative denoising process \cite{tevet2022human,zhang2024motiondiffuse}. 

While these approaches have improved motion realism and semantic control, they primarily operate on abstract modalities like text \cite{guo2024momask,guo2022generating} or music \cite{siyao2022bailando,tseng2023edge}.
Parallel efforts incorporate environmental context using explicit 3D scene representation \cite{wang2022humanise,yi2024generating,wang2024move}. However, such structured inputs are often unavailable in real-world egocentric scenarios. Crucially, both lines of work overlook \textit{egocentric visual perception}, which provides rich affordances and spatial constraints directly from first-person video. By neglecting this perceptual grounding, current models \cite{tanaka2023role,liang2024intergen} inherently decouple sensing from action. This creates a representational gap that ultimately impedes their deployment in dynamic interactive settings.

\begin{figure*}[t]
\centering
\captionsetup{font={small}}
\includegraphics[width=0.9\linewidth]{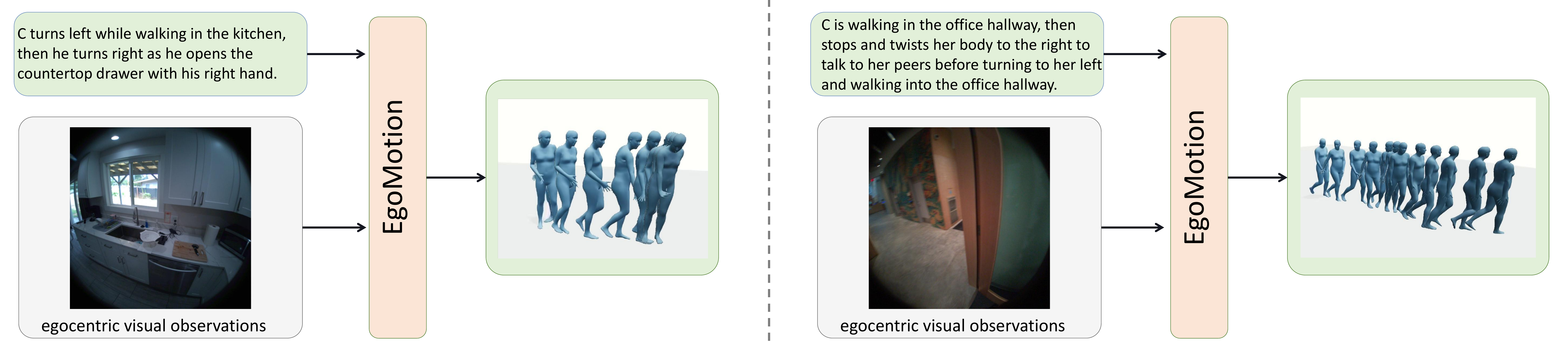}
\caption{We propose a hierarchical framework \textbf{EgoMotion} for Ego-VL Motion Generation, aiming to synthesize motion sequences conditioned jointly on egocentric visual observations and natural language instructions.}
\label{fig1}
\vspace{-10pt}
\end{figure*}

 To bridge this gap,  we introduce Egocentric Vision-Language (Ego-VL) Motion Generation. As shown in  Fig. \ref{fig1}, this task requires synthesizing motion sequences conditioned jointly on egocentric visual observations and natural language instructions. This formulation emulates real-world embodied interaction, where an agent must concurrently parse verbal commands, reason about environmental affordances, and execute feasible trajectories. 
Despite its conceptual promise, Ego-VL Motion Generation poses two major technical challenges: 

\textbf{Challenge1: Reasoning-Generation Entanglement}. A natural approach to this task is to jointly optimize a vision-language model (VLM) and a motion generator in an end-to-end fashion. This assumes the unified model can simultaneously acquire multimodal semantic understanding and precise kinematic modeling capabilities. However, such a design imposes fundamentally incompatible optimization objectives on shared model parameters. Specifically, gradients from the motion generation objective propagate into the VLM backbone, corrupting its semantic representations and undermining multimodal grounding fidelity. Conversely, the semantic reasoning objective interferes with the kinematic precision required for high-quality synthesis. This gradient interference makes end-to-end optimization a fundamentally suboptimal training paradigm for this task. 

\textbf{Challenge2: Motion Prediction Instability}. Directly regressing motion sequences in raw joint space poses a fundamental obstacle to stable, physically grounded synthesis. This raw joint space is characterized by complex biomechanical dependencies across anatomical degrees of freedom, which are notoriously difficult to capture through direct supervised regression. 
Without structural priors, a generative model must recover the underlying motion distribution across all degrees of freedom. This creates a severely underconstrained problem with no inherent guarantees of kinematic feasibility. Consequently, this unconstrained formulation makes the synthesis process highly susceptible to kinematic violations, often resulting in high-frequency joint flickering.

To address these challenges, we propose \textbf{EgoMotion}, a hierarchical framework inspired by the biological decoupling of high-level cognition and low-level motor control. EgoMotion decomposes the Ego-VL Motion Generation task into two tightly coupled yet independently optimized stages, each dedicated to a distinct objective. 

 \textbf{Stage I: Cognitive Reasoning.}  A vision-language model acts as the \textit{cognitive brain}. It performs structured multimodal reasoning to map egocentric visual observations and natural language instructions into compact motion primitives. Specifically, the VLM is trained to predict discrete motion tokens derived from a pretrained  Residual Vector Quantized Variational Autoencoder (RVQ-VAE)  \cite{lee2022autoregressive} codebook. This design reformulates motion synthesis as a symbolic sequence reasoning task, allowing the model to fully leverage the VLM's autoregressive capabilities.  Critically, by isolating this stage with a single supervised objective, EgoMotion circumvents the gradient interference common in joint end-to-end optimization. This ensures that the VLM acquires clean, uncontaminated semantic representations that are highly effective for subsequent motion generation.
 
\textbf{Stage II: Motion Generation.}   A diffusion-based motion generator acts as the \textit{motor cerebellum}. It is conditioned on the final-layer hidden states of the frozen VLM backbone, which encapsulate rich semantic context distilled from egocentric observations and language instructions. Rather than operating in raw joint space, the generator performs iterative denoising within the continuous latent space of a pretrained variational autoencoder (VAE). This design provides a well-regularized optimization landscape that substantially mitigates kinematic violations. Consequently, the generator can progressively refine coarse behavioral intents into temporally coherent and physically plausible motion trajectories.

To the best of our knowledge, this work represents one of the early systematic attempts to investigate egocentric vision-language conditioned human motion generation within a unified generative framework. Extensive experiments demonstrate that EgoMotion consistently outperforms existing baselines, yielding motion sequences that are semantically grounded and temporally coherent.
\section{Related Work}
\noindent
\subsection{Conditional Motion Generation} \ 
Conditional motion generation aims to synthesize diverse and high-fidelity motion sequences guided by external signals \cite{xue2025human,zhu2023human}, including text  \cite{cen2024generating,guo2022generating,guo2024momask}, action \cite{petrovich2021action,guo2020action2motion,aberman2020unpaired, mason2022real,liu2022investigating}, music \cite{siyao2022bailando,tseng2023edge} and 3D scenes \cite{wang2022humanise,yi2024generating,wang2024move}. Fueled by the avaliablity of large-scale motion capture datasets and rapid advances in deep generative models, this field has witnessed substantial progress in recent years. Existing approaches can be broadly categorized into three main families:  autoregressive models \cite{zhang2023generating,lu2023humantomato}, generative masked modeling \cite{pinyoanuntapong2024mmm,guo2024momask}, and diffusion-based methods \cite{zhang2024motiondiffuse,tevet2022human,raab2024monkey,kapon2024mas,shafir2023human,xu2024regennet}.

\textbf{Autoregressive approaches} typically quantize continuous motions into discrete tokens,  modeling the resulting sequence through step-by-step autoregressive prediction \cite{zhang2023generating,lu2023humantomato,jiang2023motiongpt,zhang2024motiongpt}. A representative work, T2M-GPT \cite{zhang2023generating}, adopts a Vector Quantized Variational Autoencoder (VQ-VAE)  \cite{van2017neural} with a GPT-style transformer \cite{vaswani2017attention} to perform conditional motion synthesis within a structured codebook space. Building upon this paradigm, HumanTOMATO \cite{lu2023humantomato} further incorporates a hierarchical VQ-VAE alongside GPT-based decoding modules, enabling the joint modeling of fine-grained motion dynamics across both body and hand articulations.  
Despite their promising performance, autoregressive approaches are subject to inherent architectural limitations. First, the discretization process induced by VQ-VAE \cite{van2017neural} inevitably incurs quantization errors, resulting in the loss of high-frequency kinematic details. Second, the causal attention mechanism underlying autoregressive modeling precludes temporal lookahead, restricting each token to a strictly unidirectional context. This sequential dependency renders the generation process susceptible to error accumulation, which can precipitate temporal drift and undermine the long-term coherence and fluidity of the synthesized trajectories.

\textbf{Generative masked modeling} offers a non-autoregressive alternative by reconstructing complete motion sequences from partially masked token representations  \cite{ghosh2025duetgen,pinyoanuntapong2024mmm,guo2024momask}.  For instance, MMM \cite{pinyoanuntapong2024mmm} employs a conditioned masked transformer that predicts motion tokens in parallel under the guidance of textual embeddings. 
MoMask \cite{guo2024momask} further enhances synthesis fidelity by coupling Residual Vector Quantization  \cite{lee2022autoregressive} with a residual transformer architecture. Extending this paradigm to complex multi-person scenarios, DuetGen \cite{ghosh2025duetgen} introduces a hierarchical masked modeling framework that facilitates synchronized motion synthesis between interacting dancers under musical guidance.
In contrast to autoregressive counterparts, masked generative models leverage bidirectional self-attention, enabling each token to attend to both preceding and succeeding contexts during generation. This global receptive field effectively mitigates the temporal drift inherent in sequential prediction. Nevertheless, the performance of these models still remains fundamentally  constrained by the discretization fidelity of the underlying codebook. 

\textbf{Diffusion-based methods} synthesize motion sequences by iteratively reversing a learned corruption process, progressively denoising a Gaussian prior into coherent motion trajectories.  In contrast to discrete generative paradigms, these approaches operate in continuous spaces, thereby circumventing the precision loss introduced by vector quantization. For instance, MDM \cite{tevet2022human} employs a transformer-based diffusion backbone conditioned on semantic features extracted from a pre-trained CLIP encoder \cite{radford2021learning}.  MotionDiffuse \cite{zhang2024motiondiffuse} extends this line by enabling fine-grained, part-level control through detailed textual prompts.  More recently, several works \cite{raab2024monkey,kapon2024mas,raab2023single} have demonstrated strong generalization capability across out-of-domain motion distributions.
Despite their capacity for high-fidelity synthesis, existing diffusion-based frameworks \cite{zhang2024motiondiffuse,tevet2022human} predominantly generate motion trajectories directly in raw joint space, which is prone to noticeable jitter and high-frequency artifacts due to the inherent step-wise stochasticity of the denoising process. This observation motivates our adoption of latent diffusion, wherein the denoising process is performed within the compressed latent space of a pretrained VAE, yielding smoother and more kinematically consistent motion synthesis. 

\vspace{0.5\baselineskip}
\noindent
\subsection{Scene-aware Motion Generation} \
Scene-conditioned motion generation aims to synthesize human motions that are both semantically meaningful and physically consistent with the surrounding environments. Existing approaches typically ground the generation process in explicit 3D geometric priors, including point clouds \cite{araujo2023circle, wang2022humanise, zheng2022gimo}, meshes \cite{wang2021synthesizing}, voxel grids \cite{cen2024generating, jiang2024scaling}, and signed distance fields \cite{wang2024move, yi2024generating, zhang2020place}. These works typically leverage such representations to model human-scene \cite{araujo2023circle, hassan2019resolving, jiang2024scaling} and human-object \cite{corona2020context, kulkarni2023nifty, li2024controllable, zhang2022couch} interactions.  However, acquiring high-fidelity 3D scene representations typically necessitates specialized capture hardware or computationally demanding offline reconstruction pipelines \cite{gu2024egolifter, pan2023aria, tschernezki2023epic}, imposing a significant practical bottleneck for real-world deployment. 
In contrast, egocentric images bypass the need for specialized hardware or offline reconstruction, offering a more ubiquitous and deployable alternative. Nevertheless,  motion synthesis grounded in such unstructured visual cues remains remarkably underexplored. In this work, we formally introduce the task of Egocentric Vision-Language Motion Generation, aiming to synthesize context-aware human motions jointly conditioned on a single egocentric frame and language instruction.

\vspace{0.5\baselineskip}
\noindent
\subsection{Egocentric Motion Reconstruction and Generation} \
Egocentric visual inputs provide a first-person perspective that is essential for understanding human-centric interactive motion. 
A series of works \cite{li2023ego, pan2025lookout, wang2024egocentric, wang2025ego4o, yi2025estimating} has focused on egocentric pose estimation, aiming to reconstruct full-body postures from the limited field of view of wearable cameras. Early methods often rely on optical flow to estimate the 3D trajectory of the ego-device, or depend exclusively on SLAM-based trajectory without considering broader scene context. Subsequent research integrates scene point clouds with egocentric image features to achieve scene-aware motion reconstruction. More recently, UniEgoMotion \cite{patel2025uniegomotion} demonstrates that rich scene context can be effectively captured from egocentric images, reducing the reliance on external sensors or pre-scanned maps.
Recent research has extended this line from passive estimation to proactive motion reaction and generation.  HERO \cite{yu2025hero} and EgoReAct \cite{zhang2025egoreact} synthesize plausible human  \textit{reactions} conditioned on egocentric videos, enabling visually grounded responses to environmental events. EgoTwin \cite{xiu2025egotwin} further introduces a ``dreaming" paradigm that jointly synthesizes human motion and egocentric observations, effectively learning a generative world model that couples first-person perception with motion dynamics. 
In contrast to existing methods that primarily focus on passive reconstruction or joint video-motion generation, we focus on Ego-VL Motion Generation task and propose a two-stage framework that explicitly decouples cognitive reasoning from continuous motion synthesis.
\section{Methodology}
\textbf{Problem Formulation.} \
We define Ego-VL Motion Generation as synthesizing a 3D human motion sequence $\mathcal{X}=\left\{X_t\right\}_{t=1}^N\in\mathbb{R}^{N\times D}$ conditioned on an egocentric RGB image $I_0\in\mathbb{R}^{H\times W\times 3}$, a natural language instruction $\mathcal{L}$, and an initial pose $X_0 \in \mathbb{R}^{D}$. Formally, our objective is to model the conditional distribution $p\left(\mathcal{X} \mid I_0, \mathcal{L}, X_0\right)$, where the generated motion is subject to temporal continuity from $X_0$ and semantic alignment with $I_0$ and $\mathcal{L}$. 


\begin{figure*}[t]
\centering
\captionsetup{font={small}}
\includegraphics[width=0.9\linewidth]{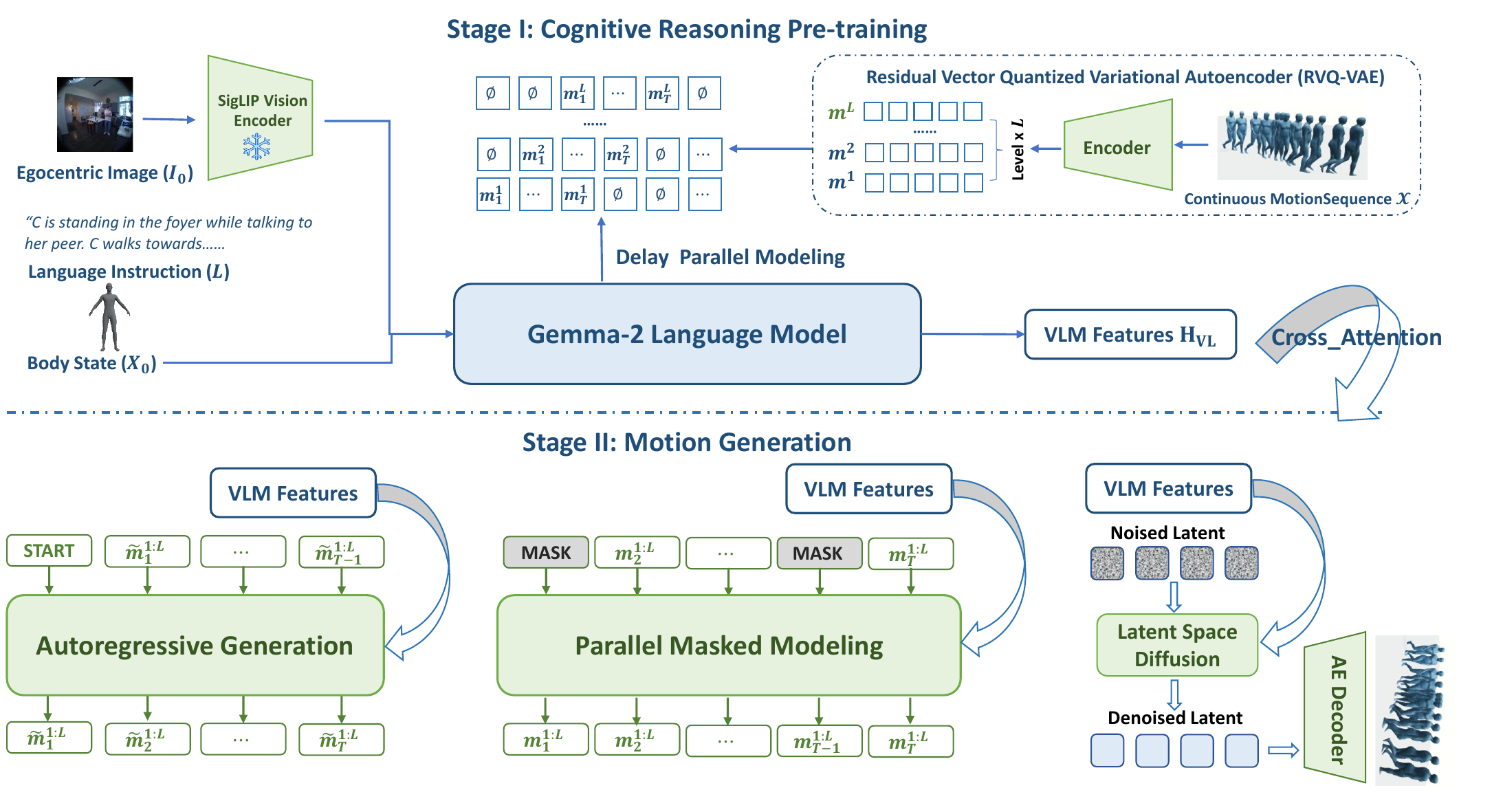}
\caption{Overview of EgoMotion. The framework employs a hierarchical two-stage paradigm. \textbf{(Stage I) Cognitive Reasoning Pre-training:} A Gemma-2 based VLM aligns multimodal inputs ($I_0, L, X_0$) with RVQ-VAE motion primitives. \textbf{(Stage II): Motion Generation}: Pre-trained VLM features ($\mathbf{H}_{\mathrm{VL}}$) condition the synthesis across three optional paradigms (Autoregressive, Masked Modeling, or Diffusion).} 
\label{fig2}
\vspace{-10pt}
\end{figure*}

\vspace{0.5\baselineskip}
\noindent
\textbf{Framework Overview.} \ 
An overview of the proposed EgoMotion framework is illustrated in Fig. \ref{fig2}. To address reasoning-generation entanglement and motion instability, EgoMotion adopts a hierarchical two-stage paradigm that mirrors the biological separation between high-level cognitive reasoning and low-level motor execution. 

\noindent
  \textbf{Stage I: Cognitive Reasoning Pre-training.} \ We first employ a Vision-Language Model (VLM) as the \textit{cognitive brain} to map egocentric observation $I_0$ and instructions $\mathcal{L}$ onto discrete motion primitives. This stage focuses on robust multimodal semantic alignment, enabling the VLM to distill high-level behavioral intent from unstructured egocentric contexts.

\noindent
  \textbf{Stage II: Latent Motor Diffusion.} \ The pretrained VLM backbone is then integrated with a diffusion-based generator, functioning as the \textit{motor cerebellum}. By conditioning on the final-layer hidden states of the VLM, a Latent Diffusion Model performs iterative denoising within a pretrained VAE's latent space. This process progressively refines noisy latent into smooth, temporally coherent 3D motion trajectories  $\mathcal{X}$, while preserving the semantic reasoning priors from Stage I. 


 \subsection{Cognitive Reasoning Pre-training Stage}
 
 The core objective of the first stage is to equip the VLM backbone with the capacity to distill high-level behavioral intentions from complex egocentric environments. Specifically, we position the VLM as a cognitive brain that establishes a structured, multimodal-to-semantic mapping, bridging the gap between raw perceptual inputs and motion primitives.

 \subsubsection{Residual Motion Tokenization}  
 Conventional discrete motion paradigms typically rely on standard Vector Quantization \cite{van2017neural} with a single, limited-capacity codebook. Such approaches inherently constrain representational power, as diverse and complex motion patterns must be compressed into a restricted set of predefined tokens. This compression bottleneck inevitably introduces significant information loss, thereby undermining the model's capacity to faithfully represent fine-grained motion nuances and limiting its expressiveness in representing the full diversity of human motion.

 Following \cite{guo2024momask}, we adopt a Residual Vector Quantized Variational Autoencoder (RVQ-VAE) \cite{lee2022autoregressive} which consists of a motion encoder $E$, a motion decoder $D$, and a set of $L$ hierarchical codebooks $\left\{B^l\right\}_{l=1}^L$. Formally, given a motion sequence $\mathcal{X}\in\mathbb{R}^{N\times D}$, the encoder $E$ maps it to a latent feature map $\mathbf{Z}\in\mathbb{R}^{N_1\times D_1}$ with a temporal downsampling ratio of $N/N_1$. To capture fine-grained dynamics while maintaining a discrete bottleneck, RVQ recursively quantizes the latent feature map $\mathbf{Z}\in\mathbb{R}^{N_1\times D_1}$. Taking $\mathbf{R}^1=\mathbf{Z}$ as the initial residual, the discrete token sequence $\boldsymbol{m}^l$ and the subsequent residual $\mathbf{R}^{l+1}$ at each quantization level $l$ are computed as:
 \begin{equation}
 \begin{split}
 &m^l_n=\boldsymbol{m}^l\left[n\right]=\arg\min_{k}\left\|\mathbf{R}^l\left[n,:\right]-e^l\left(k\right)\right\|_2^2, \\
&\mathbf{\hat{R}}^{l}=e^l\left(\boldsymbol{m}^l\right), \quad \mathbf{R}^{l+1}=\mathbf{R}^l-\mathbf{\hat{R}}^{l}.
 \end{split}
 \end{equation}
 Here, $e^l\left(k\right)$ denotes the embedding vector for index $k$ in codebook $B^l$. The final quantized representation is the summation across all levels: $\mathbf{\hat{Z}}=\sum_{l=1}^L \mathbf{\hat{R}}^{l}$. Subsequently, the decoder $D$  reconstructs the motion sequence $\hat{\mathcal{X}}=D\left(\mathbf{\hat{Z}}\right)$ from these aggregated quantized features $\mathbf{\hat{Z}}$. The model is optimized using a joint objective function:
 \begin{equation}
 \mathcal{L}_{rvq} = \left\|\mathcal{X}-\hat{\mathcal{X}}\right\|_1+\beta\sum_{l=1}^L\left\|\mathbf{R}^l-\mathrm{sg}\left[\mathbf{\hat{R}}^l\right]\right\|_2^2,
 \label{eq2}
 \end{equation}
 where $\mathrm{sg}\left[\cdot\right]$ denotes the stop-gradient operation, and $\beta$ is a hyperparameter controlling the commitment loss.

 \subsubsection{Alignment via Generative Pre-training} 
 
 \textbf{VLM Architecture.} \ 
 We adopt PaliGemma-2 \cite{steiner2024paligemma2} as the vision-language backbone. It integrates a SigLIP \cite{zhai2023sigmoid} vision encoder with a linear projection layer for cross-modal feature alignment, followed by a Gemma-2 \cite{gemma2024gemma2} language model for unified multimodal reasoning. This architecture provides a scalable and expressive foundation for learning structured motion priors conditioned on egocentric visual perception and natural language instructions.
  
  \vspace{0.5\baselineskip}
 \noindent
 \textbf{Delay Parallel Modeling.} \ 
Let  $\left\{\boldsymbol{m}^1,\dots,\boldsymbol{m}^L\right\}$ denote the hierarchical motion token streams produced by residual motion tokenization. To enable efficient multi-stream autoregressive generation, we adopt a delay parallel modeling formulation following \cite{hou2025motionverse}, which supports simultaneous prediction across all residual streams. Specifically, each stream $\boldsymbol{m}^l$ is temporally shifted to construct a delayed sequence $\boldsymbol{\tilde{m}}^l$ of length  $N_1+L-1$: 
 \begin{equation}
 \boldsymbol{\tilde{m}}^l=\left[\underbrace{\varnothing,\dots,\varnothing}_{l-1}, \boldsymbol{m}^l,\underbrace{\varnothing,\dots,\varnothing}_{L-l}\right],
 \end{equation}
 where $\varnothing$ represents a special padding token. This structured staggering allows the VLM to employ $L$ parallel prediction heads, generating tokens from all residual streams concurrently at each decoding step. Conditioned on egocentric visual observations $I_0$ and language instruction $\mathcal{L}$, the VLM backbone leverages causal self-attention to capture the joint conditional distribution at each timestep $n$: 
 \begin{equation}
 \begin{split}
 &p\left( \tilde{m}^{1:L}_n| \tilde{m}^{1:L}_{<n},I_0,\mathcal{L}, X_0\right)=\\
 &p\left( \tilde{m}^1_n,\dots,\tilde{m}^L_n| \tilde{m}^1_{<n},\dots,\tilde{m}^L_{<n},I_0,\mathcal{L}, X_0\right)= \\
  &p\left( m^1_n,\dots,m^L_{n-L+1}| m^1_{<n},\dots,m^L_{<n-L+1},I_0,\mathcal{L}, X_0\right).
 \end{split}
 \end{equation}
 By transforming hierarchical residual prediction into a temporally interleaved autoregressive process, this formulation jointly captures (i) intra-stream temporal dependencies via $m^l_{<n}$, and (ii) inter-stream residual dependencies via $m^{<l}_n$, within a single forward pass. This strategy substantially improves decoding efficiency while preserving fine-grained motion expressiveness.

  \vspace{0.5\baselineskip}
 \noindent
 \textbf{Training Objective.} \ 
 During pre-training, we optimize the VLM to maximize the log-likelihood of the delayed motion tokens via an autoregressive cross-entropy loss:
 \begin{equation}
 \mathcal{L} =\sum_{n=1}^{N_1+L-1} \log p\left( \tilde{m}^{1:L}_n| \tilde{m}^{1:L}_{<n},I_0,\mathcal{L}, X_0;\Theta\right),
 \end{equation}
 where $\Theta$ denotes the parameters of the VLM backbone. 
 This multi-stream prediction objective provides structured supervision that encourages the model to internalize motion dynamics, while simultaneously regularizing the VLM's latent representation space. These learned reasoning priors effectively bridge high-level semantic intent with fine-grained kinematic constraints, and subsequently serve as task-aware conditioning signals for the downstream motion generation stage.

 \subsection{Motion Generation Stage}
 As shown in Fig. \ref{fig2}, to systematically investigate the most effective interface between cognitive reasoning and motion generation, we instantiate three representative generative paradigms built upon a unified Conditional Transformer architecture. In particular, the motion generator employs cross-attention blocks to attend to the VLM-derived representation $\mathbf{H}_{\mathrm{VL}}$ from Stage I:
 \begin{equation}
 \mathbf{H}_{\mathrm{VL}}=\mathrm{VLM}\left(I_0,\mathcal{L}, X_0; \Theta\right).
 \end{equation}
 Within this generator, cross-attention blocks enforce semantic alignment with the pre-trained cognitive prior, while self-attention blocks capture the intrinsic temporal dependencies of the motion sequence. While sharing this structural foundation, the three paradigms diverge in their respective modeling spaces and optimization objectives.

  \vspace{0.5\baselineskip}
 \noindent
 \textbf{Autoregressive Generation.}  \
 A natural instantiation of the cognitive-to-motor mapping is to extend the sequential modeling paradigm from Stage I to an Autoregressive  Generation framework. Given the delayed multi-stream tokens $\left\{\boldsymbol{\tilde{m}}^1,\dots,\boldsymbol{\tilde{m}}^L\right\}$, a Conditional Transformer is trained to maximize the conditional log-likehood:
 \begin{equation}
 \mathcal{L}_{\mathrm{AR}}=-\sum_{n=1}^{N_1+L-1}\log p\left(\tilde{m}^{1:L}_n|\tilde{m}^{1:L}_{<n},\mathbf{H}_{\mathrm{VL}};\theta\right),
 \end{equation}
 where $\theta$ denotes the parameter of the motion generator. This paradigm is conceptually appealing due to its alignment with the pre-training stage,  enabling seamless transfer of the learned likelihood-based motion distribution. However, its strictly causal dependency structure inherently introduces exposure bias: the model is conditioned exclusively on ground-truth histories during training and thus remains unexposed to its own prediction errors. Consequently,  such errors accumulate during long-horizon inference rollouts, frequently leading to temporal drift and progressive degradation in kinematic plausibility.

   \vspace{0.5\baselineskip}
 \noindent
 \textbf{Parallel Masked Modeling.}  \
 In contrast to the strictly causal autoregressive formulation, we instantiate the cognitive-motor mapping via a Parallel Masked Modeling paradigm to enable non-autoregressive motion generation. 
Given the multi-level residual token sequence $\left\{\boldsymbol{m}^1,\dots,\boldsymbol{m}^L\right\}$ from Stage 1, the model is trained to predict tokens across all $L$ levels in parallel. To enforce robust temporal reasoning, we introduce a \textbf{structured masking unit}: once a temporal index is selected for masking, the tokens across all residual levels at that timestep are jointly masked. This design prevents the model from relying on \textit{vertical} shortcuts, \ie,  inferring masked tokens from co-located residuals at the same timestep, thereby forcing the Transformer to leverage broader temporal context. The Masked Motion Transformer reconstructs these masked tokens conditioned on the visible context and semantic prior $\mathbf{H}_{\mathrm{VL}}$:
 \begin{equation}
 \mathcal{L}_{\mathrm{PMM}}=-\frac{1}{\left|\mathcal{M}\right|}\sum_{n\in\mathcal{M}}\log p\left(m^{1:L}_n|m^{1:L}_{\setminus \mathcal{M}},\mathbf{H}_{\mathrm{VL}};\theta\right),
 \end{equation}
 where $\mathcal{M}$ denotes the set of masked temporal indices.  This temporally aligned masking strategy encourages the learning of coherent dependencies and enables high-efficiency parallel generation, effectively mitigating the exposure bias inherent in autoregressive models. Nevertheless, discrete tokenization inevitably imposes a representational bottleneck. The fine-grained kinematic details may be lost during quantization, potentially compromising reconstruction fidelity and the realism of synthesized egocentric interactions.

   \vspace{0.5\baselineskip}
 \noindent
 \textbf{Latent Space Diffusion.}  \
 To overcome the representational bottleneck of discrete tokenization, we introduce a Latent-space Diffusion paradigm, formulating motion generation as a continuous denoising process within a structured latent manifold. 
 To this end, we pre-train a Variational Autoencoder (VAE)  \cite{kingma2013auto} to map raw motion sequence $\mathcal{X}$ into a continuous latent space. Specifically, the encoder $E$ parameters a posterior distribution $q_{\phi}\left(\mathbf{Z}|\mathcal{X}\right)$, from which a latent code $\mathbf{Z}$ is sampled via the reparameterization trick. The VAE model is optimized by minizing a joint objective: a Kullback-Leibler (KL) divergence loss regularizes the posterior against a standard Gaussian prior $\mathcal{N}\left(\mathbf{0}, \mathbf{I}\right)$, while a  reconstruction loss ensures that the decoder $D$ can faithfully recover the motion sequence $\mathcal{\hat{X}}=D\left(\mathbf{Z}\right)$ from the latent manifold. This joint optimization yields a smooth, compact, and kinematically informed latent space, providing a structured inductive bias for the subsequent flow-matching process.

 Within this latent space, we employ a Flow Matching paradigm \cite{lipman2022flow}  to transport a Gaussian noise towards the target motion distribution, conditioned on the frozen semantic features $\mathbf{H}_{\mathrm{VL}}$. Specifically, given a ground-truth latent $\mathbf{Z}$, a timestep $\tau\in\left[0,1\right]$, and sampled noise $\epsilon \sim \mathcal{N}(0, I)$, the interpolated  probability path $\mathbf{Z}_\tau$ is constructed via linear interpolation:
 \begin{equation}
 \mathbf{Z}_\tau = \tau \mathbf{Z} + \left(1-\tau\right) \epsilon.
 \end{equation}
 The motion generator $f_\theta$ is trained to approximate the target velocity field $\epsilon-\mathbf{Z}$ by minimizing the conditional flow-matching objective:
 \begin{equation}
 \mathcal{L}_{\mathrm{LFM}} = \mathbb{E}_{\tau, \mathbf{Z}, \epsilon} \left[ \| f_\theta\left(\mathbf{H}_{\mathrm{VL}}, \mathbf{Z}_\tau, \tau\right) - (\epsilon - \mathbf{Z}) \|_2^2 \right].
 \end{equation}
 By operating within a continuous and compact latent space, this paradigm circumvents the quantization artifacts inherent in discrete token-based modeling while preserving fine-grained kinematic details. Furthermore, the KL-regularized latent space provides a significantly 
 smoother optimization landscape compared to raw-space diffusion, yielding enhanced training stability and superior synthesis fidelity.

\section{Experiment}
\subsection{Experimental Settings}
\textbf{Motion Representation.} \ 
To facilitate effective alignment between 3D human motion and egocentric video, we adopt a head-centric motion representation. Conventional root-centric formats \cite{guo2022generating} typically encode head pose implicitly through nested kinematic chains. This structure makes it inherently difficult for neural networks to model the precise head trajectories essential for egocentric tasks. In contrast, the head-centric representation explicitly parameterizes motion relative to the head's orientation and position. 
Specifically, for a motion sequence with $J$ joints, the pose at each timestep is represented by  a $\left(3J+8\right)$-dimensional feature vector: $\left[\mathbf{\dot{v}}_{xz}, \mathbf{r}_{\Delta}, \mathbf{p}_{\mathrm{local}}\right]$. Here, $\mathbf{\dot{v}}_{xz}\in\mathbb{R}^{2}$ denotes the horizontal root velocity on the $\mathrm{XZ}-$plane, transformed into the local coordinate frame of the previous timestep to ensure translation invariance. The term  $\mathbf{r}_{\Delta}\in\mathbb{R}^6$ captures the frame-to-frame change in global yaw via a 6D rotation representation. Finally, $\mathbf{p}_{\mathrm{local}}\in\mathbb{R}^{3J}$ encodes the 3D positions of all $J$ joints within a head-centric, heading-aligned coordinate space. By using this coordinate space, we explicitly expose the spatial relationship between the body and the observer's viewpoint, simplifying the learning process for egocentric alignment. 

\begin{table*}[t]
\centering
\caption{Quantitative results of egocentric vision-language (Ego-VL) motion generation. By default, EgoMotion is equipped with a Latent Diffusion Motion Generaton with two-stage training.}
\begin{tabular}{l c c c c c c c}
\toprule
\multirow{2}{*}{Methods} & \multicolumn{1}{c}{Fidelity} & \multicolumn{2}{c}{Semantic Alignment }  & \multicolumn{2}{c}{Physical Plausibility} & \multicolumn{2}{c}{Temporal Smoothness} \\
\cmidrule(lr){2-2} \cmidrule(lr){3-4}   \cmidrule(lr){5-6}  \cmidrule(lr){7-8} 
& FID $\downarrow$ & R@Top1 $\uparrow$ & MM-Dist $\downarrow$ $\rightarrow$ & FS $\downarrow$ & FC $\downarrow$  & Acce $\downarrow$ & Jerk $\downarrow$ \\
\midrule
\color{mygray}{Real} & - &\color{mygray}{67.4} &\color{mygray}{0.897}  &- & -& \color{mygray}{0.0015} &\color{mygray}{0.0010}  \\
T2M-GPT \cite{zhang2023generating} &0.0655 &28.1 &1.136 &1.30 &15.5 &0.0029 &0.0045 \\
MMM \cite{pinyoanuntapong2024mmm} &0.1005 &36.1 &1.104 &1.32 &15.6 &0.0024 &0.0039 \\
MoMask \cite{guo2024momask} &0.0621 &41.2 &1.125 &1.98 &17.2 &0.0017 &0.0024 \\
MotionDiffuse \cite{zhang2024motiondiffuse} &0.0609 &37.2 &1.062 &1.18 &12.4 &0.0016 &0.0020 \\
\midrule
\textbf{EgoMotion (Ours)} & \textbf{0.0018} & \textbf{69.4} & \textbf{0.872}  & \textbf{1.05} & \textbf{12.3} & \textbf{0.0015} & \textbf{0.0015} \\
\bottomrule
\end{tabular}
\label{tab1}
\end{table*}
\vspace{0.5\baselineskip}
\noindent
\textbf{Datasets.} \ 
We evaluate our model on Nymeria \cite{ma24eccv}, a large-scale, multi-modal dataset collected in unconstrained real-world environments.  Unlike  lab-captured or synthetic benchmarks, Nymeria covers a diverse range of daily activities performed by various subjects across various indoor and outdoor scenes. The dataset provides synchronized triplets of: i) Egocentric Video: Recorded via Project Aria glasses, where the field-of-view is intrinsically aligned with natural head orientation; ii) 3D Motion: full-body kinematics  captured through Xsens inertial sensors; iii) Textual Descriptions:  Fine-grained action descriptions  curated by human annotators.
For our experiments, we partition the continuous recordings into non-overlapping 5-second segments, yielding a total corpus of approximately 140K samples. We employ an $80\%/20\%$ split for the training and test sets, respectively. 

\vspace{0.5\baselineskip}
\noindent
\textbf{Evaluation Metrics.} \
We assess the generated motions across four complementary dimensions to provide a holistic evaluation of generative quality.
\begin{itemize}
\item \textbf{Motion Fidelity.} We employ the Fréchet Inception Distance (\textbf{FID}) as the primary indicator of overall motion quality. FID measures the distribution discrepancy between high-level feature spaces of the generated and ground-truth motions. 
\item \textbf{Semantic Alignment.} To evaluate how faithfully the generated motions adhere to the multi-modal conditioning, we utilize two metrics: Retrieval Precision (\textbf{R@Top1}), which quantifies the top-1 accuracy of retrieving the correct conditioning pair for a given generated motion within a batch of $64$ samples; and Multi-Modal Distance (\textbf{MM-Dist}), which measures the average Euclidean distance between motion and conditioning embeddings in a shared latent space.
\item \textbf{Physical Plausibility.} To verify adherence to basic physical constraints, we evaluate foot-ground interactions via two metrics: Foot Sliding (\textbf{FS}) \cite{he2022nemf}, which quantifies the horizontal displacement of feet during interval where they are excepted to remain stationary, serving as an indicator of the ``skating" artifact; and Foot Contact (\textbf{FC}), which measures the mean vertical distance (in millimeter) between foot joints and the ground plane, capturing ``floating" and ``penetration" artifacts.
\item \textbf{Temporal Smoothness.} To assess kinematic stability, we analyze higher-order temporal derivatives of joint positions. Acceleration (\textbf{Acce}) and \textbf{Jerk} are computed as the Euclidean norm of the second- and third-order temporal derivatives of joint positions, respectively. Lower values indicate reduced jitter and improved temporal consistency, reflecting more natural and fluid motion sequences.
\end{itemize}

To compute these metrics accurately, we develop a dedicated retrieval-based evaluation network for multi-modal egocentric alignment. The architecture consists of three specialized encoders: (i) a motion encoder comprising six transformer layers \cite{dosovitskiy2020image}; (ii) a visual encoder leveraging a pre-trained CLIP \cite{radford2021learning} visual backbone to extract egocentric first-frame appearance; and (iii) a BERT-based text encoder \cite{devlin2019bert}.
Specifically, the image and text features are concatenated and processed through four additional transformer layers to capture deep inter-modal correlations between the initial scene context and the textual instruction.  
The resulting joint conditioning embedding is then aligned with the motion representation via contrastive learning. By minimizing the distance between matched motion-conditioning pairs while maximizing separation from negatives in the shared latent space, the network learns a highly discriminative embedding suitable for evaluating cross-modal consistency.

\subsection{Implementation Details} 
\noindent
In our experiment, we adopt the 3B-parameter variant of PaliGemma-2  \cite{steiner2024paligemma2}  as the VLM backbone, with all input images resized to $224\times 224$ pixels. Motion data follows the Xsens skeletal format with $J=23$ joints, and each segment is standardized to 150 frames at 30 FPS. By default, our EgoMotion employs a latent diffusion model as the core motion generator. 

\vspace{0.5\baselineskip}
\noindent
\textbf{Motion Tokenizer and VAE Training.} \ 
For both the discrete motion tokenizer (RVQ-VAE) and the continuous VAE (employed in the latent diffusion model), we adopt a symmetric autoencoder architecture with $2\times$ temporal downsampling. The RVQ-VAE comprises $L=6$ quantization layers, each maintaining a codebook of $512$ codes. The continuous VAE compresses the raw pose sequence into a 16-dimensional latent representation.  Both models are trained using the same optimization strategy. We use the AdamW \cite{loshchilov2017decoupled} optimizer with a batch size of $256$ for 60K iterations. The learning rate is initialized at $5\times10^{-5}$ and follows a cosine annealing schedule.  For the RVQ-VAE, the commitment loss weight $\beta$ in $\mathcal{L}_{\mathrm{rvq}}$ (Eq. \ref{eq2}) is set to $0.02$.

\begin{table*}[t]
\centering
\caption{\textbf{Ablation study on key design components of EgoMotion framework.} MG denotes the motion generator, and ME refers to the motion encoder responsible for mapping motion sequences into continuous and discrete latent spaces. The symbol ``-" indicates that the model performs direct prediction in the raw joint coordinate space without latent encoding.}
\begin{tabular}{l | l l |c c c c c c c}
\toprule
\multirow{2}{*}{Methods} & \multirow{2}{*}{MG} &\multirow{2}{*}{ME} & \multicolumn{1}{c}{Fidelity} & \multicolumn{2}{c}{Semantic Alignment }  & \multicolumn{2}{c}{Physical Plausibility} & \multicolumn{2}{c}{Temporal Smoothness} \\
\cmidrule(lr){4-4} \cmidrule(lr){5-6}   \cmidrule(lr){7-8}  \cmidrule(lr){9-10} 
& & & FID $\downarrow$ & R@Top1 $\uparrow$ & MM-Dist $\downarrow$ & FS $\downarrow$ & FC $\downarrow$  & Acce $\downarrow$ & Jerk $\downarrow$ \\
\midrule
EgoMotion-StageI &None &RVQ-VAE & 0.041 & 57.9 & 0.951 &1.32  &16.8 &0.0022 &0.0030\\
\midrule
Frozen-VLM &Diffusion & -&0.030 & 63.0 & 0.916  &2.01 &13.0 & 0.0053 & 0.0094 \\
Joint-Tuning &Diffusion & -& 0.027 & \textbf{71.9} & 0.881 &1.63  &13.1 &0.0044 & 0.0076 \\
EgoMotion & Diffusion &-& \textbf{0.015} & 69.8 & \textbf{0.867}  &\textbf{1.56} &12.6 & \textbf{0.0042} & \textbf{0.0073} \\ 
\midrule
\multirow{6}{*}{EgoMotion} &Autoregressive &VQ-VAE &0.051 & 61.7 & 0.919 &1.29  &13.2 &0.0027 & 0.0041 \\
&Autoregressive &RVQ-VAE &0.028 & 60.1 &0.934 &1.15 &12.9 &0.0017 & 0.0023\\
\cmidrule(lr){2-10} 
&Masked & VQ-VAE &0.053 &62.7 &0.915 &1.27 &14.8 &0.0028 &0.0043 \\
&Masked & RVQ-VAE & 0.031 & 69.1 & 0.878 &1.22 &17.9 &0.0029 & 0.0041 \\
\cmidrule(lr){2-10}
&Diffusion & - &\textbf{0.015} & \textbf{69.8} & \textbf{0.867}  &1.56 &12.6 & 0.0042 & 0.0073 \\
&Latent Diffusion & VAE &0.018 & 69.4 & 0.872 &\textbf{1.05} &\textbf{12.3} & \textbf{0.0015} &\textbf{0.0015} \\
\bottomrule
\end{tabular}
\label{tab2}
\end{table*}

\vspace{0.5\baselineskip}
\noindent
\textbf{EgoMotion Training.} \ 
We adopt a two-stage training procedure to effectively bridge visual-linguistic reasoning with motion generation. In the first stage, we optimize the PaliGemma-2 backbone for 30K iterations using the AdamW \cite{loshchilov2017decoupled} optimizer with a constant learning rate of $1\times 10^{-4}$ and a batch size of 256. During this phase, the parameters of the vision encoder  remain frozen. This strategy preserves the integrity of the pre-trained visual representations, allowing the language model to focus exclusively on aligning visual features with the target reasoning tasks.

In the second stage, we train the motion generator for 140K iterations with a batch size of 256. We maintain a constant learning rate of $1\times 10^{-4}$ using the AdamW \cite{loshchilov2017decoupled} optimizer. To provide a stable conditioning signal, the VLM is initialized with the weights from the first stage and remains gradient-detached throughout this stage. By freezing the VLM, we ensure that the learned visual-linguistic representations remain intact, enabling the motion generator to focus solely on mapping the multi-modal latent space to high-fidelity motion sequences.

\subsection{Main Results.} 

\textbf{Baselines.} \ 
Since egocentric vision-language motion generation remains largely unexplored, direct comparisons with task-specific baselines are limited. Consequently, we evaluate our method against several state-of-the-art models from the broader text-to-motion domain, including T2M-GPT \cite{zhang2023generating},  MMM \cite{pinyoanuntapong2024mmm}, MoMask \cite{guo2024momask}, MotionDiffuse \cite{zhang2024motiondiffuse}. 
To ensure a rigorous and fair comparison, we adapted these approaches to our multi-modal egocentric setting by replacing the original text-only conditioning with a joint visual-linguistic representation. Specifically, we concatenate the egocentric image feature (extracted via a frozen SigLIP vision encoder  \cite{zhai2023sigmoid}), textual instruction embeddings, and the initial human pose embedding. This fused representation is then fed into the respective generative backbones as the conditioning signal. All adapted baselines are retrained on the Nymeria dataset using their optimal hyperparameter configurations to ensure peak performance.

\vspace{0.5\baselineskip}
\noindent
\textbf{Results.} \ Tab. \ref{tab1} evaluates our method against four state-of-the-art baselines on the Nymeria dataset. EgoMotion consistently outperforms all adapted baselines across all evaluated dimensions. Our method achieves a significant breakthrough in motion quality, with an FID of 0.0018, representing a nearly 30-fold improvement over the strongest baseline MotionDiffuse (0.0609).  Our approach also demonstrates superior temporal stability and physical grounding. EgoMotion attains the lowest acceleration and jerk, effectively matching the smoothness of ground-truth recordings. Furthermore, the marked reduction in foot sliding and foot contact errors highlights our model's ability to synthesize physically plausible sequences. 

\begin{table}[t]
\centering
\caption{\textbf{Training and Inference Efficiency of Different Model Variants.} \textbf{Training Time (TT)}: total wall-clock time to convergence (hours); \textbf{Inference Time (IT)}: mean per-sample latency at batch size 1 (milliseconds).}
\begin{tabular}{l | l l |c c}
\toprule
Methods & MG & ME & TT (h) & IT (ms) \\
\midrule
Frozen-VLM &Diffusion & -&\textbf{40.8} & 295 \\
Joint-Tuning &Diffusion & -&202.6 & 295 \\ 
EgoMotion &Diffusion & - &112.4 & 295\\
\midrule
EgoMotion-StageI &None &RVQ-VAE & 62.5 &4846\\
\midrule
\multirow{6}{*}{\shortstack[c]{EgoMotion}} &Autoregressive &VQ-VAE &+79.3 &895\\
&Autoregressive &RVQ-VAE &+81.6 &987 \\
&Masked & VQ-VAE &+61.6 & 295\\
&Masked & RVQ-VAE &+65.3 & 301\\
&Diffusion & - &+49.9 & 295\\
&Latent Diffusion & VAE &+40.8 & \textbf{226} \\
\bottomrule
\end{tabular}
\label{tab4}
\end{table}

\subsection{Ablation Studies}
This section presents a comprehensive ablation study to evaluate the key design components of EgoMotion. 

\begin{table*}[t]
\centering
\caption{\textbf{Ablation study on multi-modal conditions of EgoMotion.} All variants are equipped with a Latent Diffusion Motion Generator. The symbols Visual ($\checkmark$) and Language ($\checkmark$) indicate the inclusion of egocentric image and textual instructions as control signals, respectively.}
\begin{tabular}{l | c c |c c c c c c c}
\toprule
\multirow{2}{*}{Methods} & \multirow{2}{*}{Visual} &\multirow{2}{*}{Language} & \multicolumn{1}{c}{Fidelity} & \multicolumn{2}{c}{Semantic Alignment }  & \multicolumn{2}{c}{Physical Plausibility} & \multicolumn{2}{c}{Temporal Smoothness} \\
\cmidrule(lr){4-4} \cmidrule(lr){5-6} \cmidrule(lr){7-8}  \cmidrule(lr){9-10} 
& & & FID $\downarrow$ & R@Top1 $\uparrow$ & MM-Dist $\downarrow$ & FS $\downarrow$ & FC $\downarrow$  & Acce $\downarrow$ & Jerk $\downarrow$ \\
\midrule
\multirow{3}{*}{EgoMotion} &$\checkmark$ & $\times$  &0.041 & 30.0 & 1.122 &1.08  &12.5 &\textbf{0.0014} &\textbf{0.0014} \\
 &$\times$    &$\checkmark$  &0.022 & 67.4 & 0.874 & 1.25 &13.1 & 0.0015 & 0.0015\\
  &$\checkmark$    &$\checkmark$  &\textbf{0.018} & \textbf{69.4} & \textbf{0.872} &\textbf{1.05} & \textbf{12.3}& 0.0015 &0.0015\\
\bottomrule
\end{tabular}
\label{tab3}
\end{table*}

\subsubsection{Effectiveness of Two-Stage Training Strategy} \ 
We evaluate the proposed two-stage training strategy by comparing EgoMotion with two representative variants: Frozen-VLM and Joint-Tuning.  Frozen-VLM fixes the pre-trained VLM as a static feature extractor and trains only the motion generator, and Joint-Tuning performs end-to-end optimization of  VLM and motion generator jointly. 

\textbf{(i) Frozen-VLM.} As shown in Tab. \ref{tab2}, Frozen-VLM yields the weakest performance across nearly all metrics (\eg, FID of 0.030 and FS of 2.01). This indicates that frozen VLM features lack the domain-specific knowledge required for precise egocentric motion synthesis. This deficiency manifests as both degraded generation fidelity and pronounced physical artifacts such as significant foot sliding. 
 
\textbf{(ii) Joint-Tuning.}  While Joint-Tuning achieves the highest semantic alignment, it exhibits inferior generation fidelity (FID of 0.027) and higher physical instability (FS of 1.63) compared to EgoMotion (FID of 0.015, FS of 1.56). This performance gap indicates that simultaneous optimization may introduce gradient interference between semantic reasoning and kinematic synthesis objectives, leading to a suboptimal trade-off. Furthermore, as shown in Tab. \ref{tab4}, Joint-Tuning incurs significantly higher training costs due to large-scale parameter updates on VLM backbone, yet still underperforms EgoMotion in overall motion realism. 

In contrast, EgoMotion adopts a two-stage training strategy that explicitly decouples VLM adaptation from generative refinement. This approach effectively bridges the modality gap while fully preserving the generative capacity of the diffusion backbone. 
This principled decoupling yields a more balanced performance profile: i) Superior Quality: achieving a better trade-off across generation fidelity, semantic alignment, and physical plausibility; ii) Enhanced Efficiency: Substantially optimizes resource usage, reducing total training time by nearly 45\% (as detailed in Tab. \ref{tab4}).

\subsubsection{Effectiveness of Residual Quantization} 
We evaluate motion quantization strategies by comparing VQ-VAE and RVQ-VAE across different generative backbones. 

\textbf{(i) Autoregressive Models.} As shown in Tab. \ref{tab2}, replacing VQ-VAE with RVQ-VAE in the autoregressive generator yields substantial gains in generation fidelity. Specifically, the FID drops from 0.051 to 0.028, while motion stability improves as evidenced by a reduction in jerk from 0.0041 to 0.0023. As reported in Tab. \ref{tab4}, these gains are achieved with minimal computational overhead, incuring only a marginal increase in training time (+81.6h vs. +79.3h) and comparable inference latency. 
These results suggest that the hierarchical codebooks in RVQ-VAE provide finer-grained motion representations, which are critical for accurate next-token prediction in autoregressive generation. 

\textbf{(ii) Masked Transformers.} A consistent trend is observed for masked generative models. Adopting RVQ-VAE substantially improves semantic alignment, with R@Top1 increasing from 62.7\% to 69.1\%. As shown in Tab. \ref{tab4}, this improvement is highly cost-efficient, introducing negligible inference overhead (301ms vs. 295ms).
This indicates that reduced quantization error enables more precise modeling of the correspondence between egocentric instructions and discrete motion tokens, without compromising the efficiency of the masked Transformer architecture.

Overall, RVQ-VAE consistently outperforms VQ-VAE across all evaluated generation paradigms with minimal  overhead, demonstrating its superiority as a discrete motion representation for egocentric motion synthesis.

\subsubsection{Analysis of Motion Generator} \
We evaluate three generative paradigms within the Stage II framework: Autoregressive, Masked Modeling, and Diffusion.  As shown in  Tab. \ref{tab2}, the diffusion-based generator achieves the best generation fidelity (FID of 0.015), significantly outperforming the autoregressive (0.028) and masked (0.031) variants. This gap demonstrates that diffusion models more accurately capture the continuous and multimodal nature of egocentric motion compared to discrete token-based approaches. Regarding semantic alignment, the diffusion and masked generator achieve competitive performance (R@Top1  of 69.8\% and 69.1\%), while the autoregressive model lags considerably behind (60.1\%).  This performance gap is primarily attributable to exposure bias and error accumulation in sequential decoding: early-stage prediction errors propagate through the sequence, inducing trajectory drift and semantic misalignment. In contrast, diffusion and masked models benefit from global context modeling, enabling more coherent grounding of complex instructions throughout the generated motion. 

For Inference Efficiency, the autoregressive model incurs the highest inference latency (987 ms) due to its strictly sequential decoding procedure, as shown in Tab. \ref{tab4}. Diffusion and masked models are substantially faster, both operating at approximately 300ms per sample. Notably, the Latent Diffusion Model achieves the lowest inference latency (226 ms) by performing denoising within a compressed latent space rather than the high-dimensional raw joint space. 
Overall, the diffusion-based generator offers the most favorable performance profile across all evaluated dimensions, including fidelity, semantic alignment, physical plausibility, and inference efficiency. These results establish it as the preferred motion generator within the EgoMotion framework, providing a robust foundation for high-fidelity egocentric motion synthesis.

\subsubsection{Impact of Motion Latent Spaces} 
We further investigate the impact of the VAE-based latent space on the diffusion generator.  As shown in Tab. \ref{tab2}, while the raw-space diffusion variant achieves competitive generation fidelity, it lacks inherent structural priors to ensure motion stability, leading to pronounced jitter and physical artifacts (FS of 1.56). In contrast, the latent diffusion generator markedly enhances physical plausibility by operating within a regularized, low-dimensional latent space. The VAE effectively suppresses high-frequency non-physical oscillations, achieving superior foot-sliding performance (FS of 1.05) and the most spatiotemporally coherent trajectories among all variants, with acceleration and jerk both minimized to 0.0015. These results demonstrate that the learned latent space acts as a strong structural regularizer, which is indispensable for synthesizing physically grounded and perceptually natural egocentric motion. 

Beyond motion quality, the latent diffusion generator offers substantial computational advantages. By shifting the iterative denoising process from the high-dimensional raw joint space to a compact latent representation, it achieves the lowest inference latency (226ms), a 23\% reduction compared to the raw-space baseline, as shown in Tab. \ref{tab4}. This efficiency gain proves that the VAE-based architecture not only refines the motion manifold but also significantly lowers the deployment barrier for real-time egocentric applications.

\subsubsection{Influence of Input Conditioning} 
We investigate the contribution of each input modality by ablating the conditioning signals provided to EgoMotion. Specifically, we compare the full framework against two variants: \textbf{(i) Visual-only}, which conditions the model solely on the egocentric visual observation $I_0$, removing all language guidance; \textbf{(ii) Language-only}, which conditions the generator solely on the natural language instruction $L$ while discarding the egocentric visual observation.

As shown in Tab. \ref{tab3},  the full model consistently outperforms both single-modality baselines across nearly all metrics, highlighting a powerful synergy between visual grounding and linguistic intent. 
As observed, the Visual-only variant exhibits the weakest semantic alignment, confirming that a single egocentric frame provides insufficient cues to infer high-level action intent. However, it maintains a competitive foot-sliding (FS) score of 1.08. This suggests that the visual modality primarily functions as a spatial anchor, providing the environmental constraints necessary to ground motion within the physical world. Without these visual cues, the FS score of Language-only variant degrades to $1.25$, which indicates that linguistic instructions alone cannot ensure precise kinematic alignment with the underlying 3D scene.

The Language-only variant significantly improves semantic metrics, yet it suffers from lower generation fidelity (FID of 0.022 vs. 0.018). This performance gap suggests that while language provides the global structure of the action, it lacks the fine-grained geometric cues necessary for high-fidelity synthesis. The full EgoMotion model bridges this gap, achieving the best FID (0.018) and R@Top1 (69.4\%). These results demonstrate that VLM-derived representations successfully fuse abstract textual commands with egocentric visual priors, yielding motion that is both semantically precise and physically consistent.

\begin{figure*}[t]
\centering
\captionsetup{font={small}}
\includegraphics[width=0.9\linewidth]{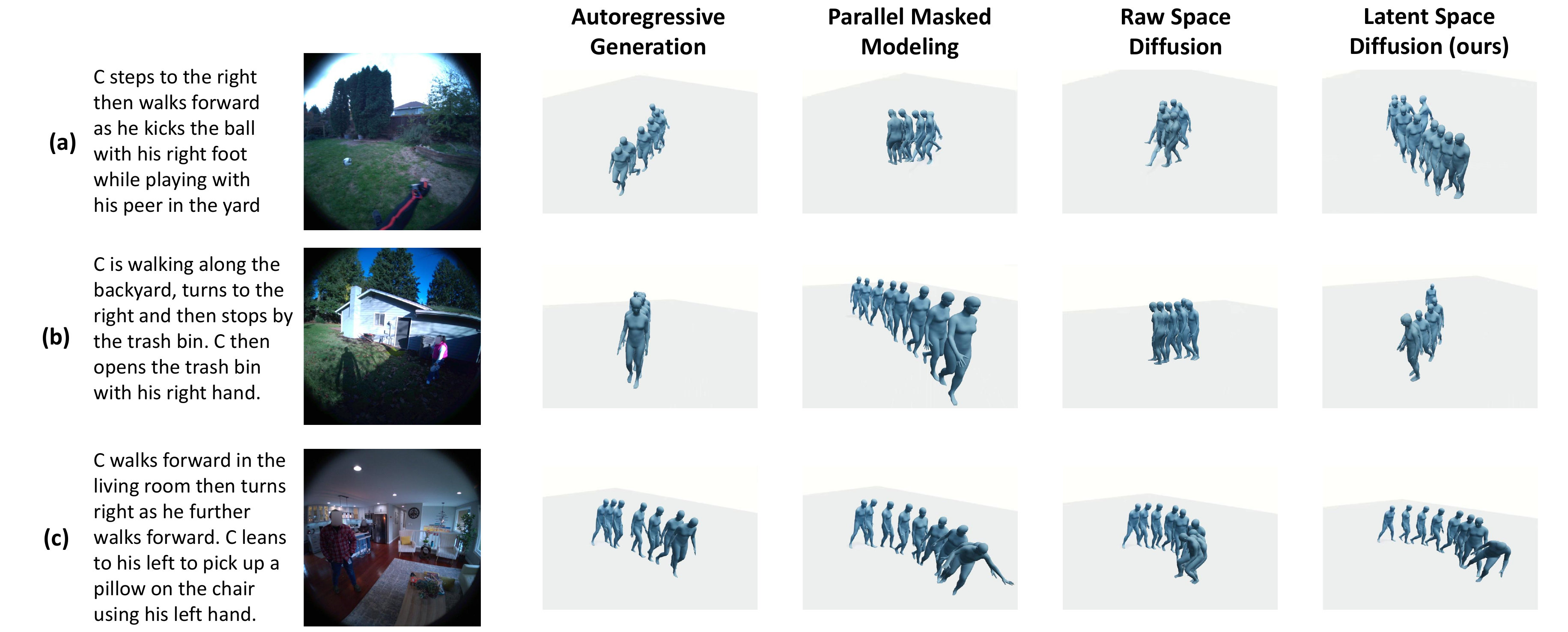}
\caption{Qualitative comparison of motion generation across different paradigms.  Generated motion sequences are visualized from a first-person perspective, conditioned on the input egocentric frame and corresponding language instructions.}
\label{fig3}
\vspace{-10pt}
\end{figure*}

\subsection{Visualization Analysis}
To qualitatively assess generation performance, we visualize motion sequences produced by the four paradigms under identical linguistic instructions across three representative scenarios in Fig. \ref{fig3}.

\vspace{0.5\baselineskip}
\noindent
\textbf{Autoregressive (AR) Generation.} \ The AR paradigm exhibits the most pronounced performance degradation, particularly in long-horizon generation tasks. 
As shown in Fig.~\ref{fig3} (b), the AR model initially captures the locomotion intent but subsequently collapses into a static or jittery state, failing to execute the ensuing ``turning'' and ``opening'' sub-actions. This observation provides direct visual evidence of exposure bias: prediction errors incurred at early timesteps accumulate and propagate through the autoregressive chain, resulting in severe temporal drift and a progressive loss 
of semantic alignment as the sequence unfolds.

\vspace{0.5\baselineskip}
\noindent
\textbf{Parallel Masked Modeling (PMM).} \ Benefiting from non-causal bidirectional attention, the PMM paradigm maintains comparatively superior global trajectory coherence, as evidenced by the walking path in Fig. \ref{fig3} (a) (kicking a ball).  Nevertheless, close inspection of the bending motion in Fig. \ref{fig3} (c) reveals conspicuous \textit{quantization artifacts}: transitions between skeletal poses appear rigid and mechanically discrete rather than smooth and fluid. This behavior exposes the representational bottleneck inherent in discrete tokenization,  which fails to preserve the fine-grained kinematic nuances critical for naturalistic motion synthesis.

\vspace{0.5\baselineskip}
\noindent
\textbf{Diffusion-Based Generation.} As shown in Fig. \ref{fig3}, both diffusion-based paradigms consistently outperform their discrete counterparts.  While \textbf{Raw Space Diffusion} produces plausible global trajectories in Fig. \ref{fig3} (a), it occasionally exhibits imprecise joint coordination during complex multi-limb interactions. This suggests that direct denoising in the high-dimensional raw joint space introduces optimization difficulties for capturing fine-grained articulation. 
In contrast, the \textbf{Latent Space Diffusion} generator synthesizes the most high-fidelity and 
semantically grounded motion sequences across all cases. As demonstrated by the dynamic ``step-and-kick'' sequence in Fig. \ref{fig3} (a) and the complex ``turn-and-lean'' sequence in Fig. \ref{fig3} (c), the latent-based approach captures intricate whole-body coordination and skeletal nuances.

 \vspace{0.5\baselineskip}
\noindent
These qualitative results visually corroborate that performing denoising within a continuous VAE latent space yields a smoother optimization landscape. By effectively circumventing the quantization loss inherent in discrete modeling, our framework facilitates the generation of highly realistic egocentric motion sequence.

\section{Conclusion}
In this work, we present EgoMotion, a hierarchical  framework for egocentric vision-language conditioned human motion synthesis. To address the  challenges of reasoning-generation entanglement, EgoMotion decouples high-level semantic reasoning from low-level kinematic synthesis via a two-stage optimization strategy. Firstly, a vision-language  module maps multimodal inputs into structured motion primitives. Second,  a latent diffusion generator synthesizes physically plausible trajectories through iterative denoising within a regularized VAE latent space. Extensive evaluations and ablation studies demonstrate that EgoMotion achieves state-of-the-art performance and validate the efficacy of its components. 

\vspace{0.5\baselineskip}
\noindent\textbf{Limitations.} While EgoMotion demonstrates strong performance in egocentric motion generation, the current framework is limited to the synthesis of human motion sequences. A  promising direction is to bridge the generated motion with physical execution.  For instance,  the generated trajectories could be used to drive humanoid robots or embodied agents in both simulated and real-world  environments. We leave the integration of EgoMotion with embodied control pipelines as an important direction for future work.


\ifCLASSOPTIONcompsoc
  \section*{Acknowledgments}
\else
  \section*{Acknowledgment}
\fi

This work is partially supported by National Natural Science Foundation of China (NSFC): 62376259 and 62306301.

\ifCLASSOPTIONcaptionsoff
  \newpage
\fi

\bibliographystyle{plain}
\bibliography{example_paper}

\end{document}